\documentclass[11pt]{article}
\usepackage{fancyhdr} % Required for custom headers
\usepackage{lastpage} % Required to determine the last page for the footer
\usepackage{extramarks} % Required for headers and footers
\usepackage[usenames,dvipsnames]{color} % Required for custom colors
\usepackage{graphicx} % Required to insert images
\usepackage{listings} % Required for insertion of code
\usepackage{courier} % Required for the courier font
\usepackage{amssymb,amsmath}
\usepackage{amsfonts}
\usepackage{mathtools}
\usepackage{subfigure}
\usepackage{enumitem}
\usepackage{bm}
\usepackage{url}
\usepackage[stable]{footmisc}
\usepackage{booktabs}
\usepackage{amsthm}
\usepackage[square]{natbib}
\usepackage{indentfirst}
\usepackage{hyperref}

\usepackage{multicol}
\usepackage{mymath}
\title{
\textbf{Rademacher Complexity of the Restricted Boltzmann Machine} \\ \textsc{asymptotic condition and CD-1 approximation} \\ 
\normalsize\vspace{0.1in}
\today \\
}

\author{
	\textbf{Xiao (Cosmo) Zhang} \\
	Department of Computer Science\\
	\texttt{zhang923@purdue.edu}
}
\date{}

\begin{document}

\maketitle
\begin{abstract}
Boltzmann machine, as a fundamental construction block of deep belief network and deep Boltzmann machines, is widely used in deep learning community and great success has been achieved. However, theoretical understanding of many aspects of it is still far from clear. In this paper, we studied the Rademacher complexity of both the asymptotic restricted Boltzmann machine and the practical implementation with single-step contrastive divergence (CD-1) procedure. Our results disclose the fact that practical implementation training procedure indeed increased the Rademacher complexity of restricted Boltzmann machines. A further research direction might be the investigation of the VC dimension of a compositional function used in the CD-1 procedure.
\end{abstract}

\section{Introduction}
A restricted Boltzmann machine (RBM) is a generative graphical model that can learn a probability distribution over its set of inputs. Initially proposed by \cite{Smolensky1986} for modeling cognitive process, it grew to prominence after successful application were found by Geoffrey Hinton and his collaborators \cite[]{Hinton2006a, Hinton2012, Salakhutdinov2009}. As a building block for deep belief network (DBN) and deep Boltzmann machines (DBM), RBM is extremely useful for pre-training the data by projecting them to a hidden layer. Also, it is proved that by adding another layer on top of a RBM, the variational lower bound of the data likelihood can be increased \cite[]{Hinton2006b, Salakhutdinov2012}, which conveys the theoretical advantage of building multilayer RBMs.

Pre-training of the data by using a RBM is essentially a unsupervised learning process, in which no label of the data is provided. Instead, the training process is trying to maximize the data likelihood by finding a proper set of parameters of the RBM.

However less attention has been given to the analysis of Rademacher complexity on RBMs. Rademacher complexity in the computational learning theory, measures richness of a class of real-valued functions with respect to a probability distribution. It can be regarded as a generalization of PAC-Bayes analysis. Its particular setting can help analysis of unsupervised learning algorithms, rather than merely the prediction problems, given the hypothesis class is possibly infinite. \cite{honorio2012} also proved that discrete factor graphs, including Markov random fields, are Lipschitz continuous, which motivated this work to further investigate the properties of RBM.

The goal of this paper is trying to bound the Rademacher complexity for the likelihood of the RBM algorithm from a given training data set, with pre-assumptions that the model structure of the RBM is known (data dimensionality and number of hidden nodes).

\section{Preliminaries}
In the beginning of this section we introduce Lipschitz continuity.

\paragraph{Definition 1.}\em 
Given the parameters $ \mathbf{\Theta} \in \mathbb{R}^{M_{1}\times M_{2}}, $ a differentiable function $ f(\mathbf{\Theta}) \in \mathbb{R}$ is called Lipschitz continuous with respect to the $ l_{p} $-norm of $  \mathbf{\Theta}, $ if there exist a constant $ K \ge 0 $ such that: 
\begin{equation}
(\forall  \mathbf{\Theta}_{1},  \mathbf{\Theta}_{2}) |f( \mathbf{\Theta}_{1}) - f( \mathbf{\Theta}_{2})| \le K\norm{ \mathbf{\Theta}_{1} -  \mathbf{\Theta}_{2}}
\end{equation} or equivalently: 
\begin{equation}
(\forall \mathbf{\Theta}) \norm{\dfrac{\partial f}{\partial \mathbf{\Theta}}} \le K
\end{equation}\em

Next we introduce the Rademacher complexity.

\paragraph{Definition 2.}\em
\subparagraph{Definition 2.1.} 
A random variable $ x \in \{-1, +1\} $  is called Rademacher  $ \iff \mathbb{P}(x) \sim Bernoulli(0.5). $
\subparagraph{Definition 2.2.}
The empirical Rademacher complexity of the hypothesis class $ \mathcal{H} $ w.r.t a data set $ \mathcal{S} = \{z^{(1)}\dots z^{(n)}\} $ is defined as:
\begin{equation}
\hat{\mathfrak{R}}_{s}(\mathcal{F}) = \mathbb{E}_{\sigma}\left[\sup_{h \in \mathcal{H}} \left(\dfrac{1}{n}\sum\limits_{i=1}^{n}\sigma_{i}h(z^{(i)})\right)\right]
\end{equation}\em

At the end of this section we give formal definition of the restricted Boltzmann machine.

\paragraph{Definition 3.}\em 
The restricted Boltzmann machine is a two layer Markov Random Field, where the observed binary stochastic visible units $ \x \in \{0, 1\}^{k} $ have pairwise connections to the binary stochastic hidden units $ \h \in \{0, 1\}^{m}. $ There are no pairwise connections within the visible units, nor within the hidden ones. Restricted Boltzmann machine is a energy-based model, in which we define the energy for a state $ \{\x, \h\} $ as 
\begin{equation}
Energy(\x, \h; \btheta) = -\x\Tr\b  - \h\Tr\c - \x\Tr\W\h,
\end{equation} where $ \btheta =  \{\c, \b, \W\}, \c \in \mathbb{R}^{m}, \b \in \mathbb{R}^{k}, $ and $ \W \in \mathbb{R}^{k \times m}. $ Hence, we can write the likelihood for an observation $ \x $ as 
\begin{equation}
p_{\btheta}(\x) = \dfrac{\sum\limits_{\h}\exp\{-Energy(\x, \h; \btheta)\}}{Z_{\btheta}},
\end{equation}
where 
\begin{equation}
Z_{\btheta} = \sum\limits_{\h}\sum\limits_{\x}\exp\{-Energy(\x, \h; \btheta)\},
\end{equation}
which is the partition function, used for normalization.
The sum over $ \h $ and $ \x $ enumerate all the possible values for the visible units and the hidden ones. Our goal optimization is to maximize the log-likelihood (minimize the negative log-likelihood) of the model. For $ N $ data samples, we can write the log-likelihood as: 
\begin{equation}
\ln p_{\btheta}(\x) = \underbrace{\ln \{\sum\limits_{\h}\exp\{-Energy(\x, \h; \btheta)\}\}}_{1} - \underbrace{\ln Z_{\btheta}}_{2} \label{LL}
\end{equation} \em

\section{Rademacher Complexity}
In this section, we provides a up-bound of the empirical Rademacher complexity for the likelihood of the restricted Boltzmann machine. Since part 2, the partition function of the restricted Boltzmann machine, of equation \ref{LL} is not depending on the data set. This part does not have any randomness and the Rademacher complexity of it is $ 0 $ by the definition. Thus, we can only focus on the Rademacher complexity of part 1 of equation \ref{LL}. Denote $ \W_{j} $ as the j-th column of the matrix $ \W, c_{j} $ as the j-th element of $ \c, h_{j} $ as the j-th element of $ \h. $ By expanding part 1 of equation \ref{LL}, we get
\begin{align}
part\, 1&=\ln \{\sum\limits_{\h}\exp\{-Energy(\x, \h; \btheta)\}\}\\
&=\ln \{\sum\limits_{h_{1}}\dots\sum\limits_{h_{m}}\exp\left[\x\Tr\b + \sum\limits_{j=1}^{m}\x\Tr\W_{j}h_{j} + \sum\limits_{j=1}^{m}h_{j}c_{j}\right]\}\\
&=\ln \{\prod\limits_{j=1}^{m}\left[\sum\limits_{h_{j} \in \{0, 1\}}\exp \left( \x\Tr\b + \sum\limits_{j=1}^{m}\x\Tr\W_{j}h_{j} + \sum\limits_{j=1}^{m}h_{j}c_{j}\right)\right]\} \label{fac}\\
&=\sum\limits_{j=1}^{m} \ln\left[\exp(\x\Tr\b) + \exp(\x\Tr\b + \x\Tr\W_{j} + c_{j} ) \right] \label{decoupled}
\end{align}

\paragraph{Lemma 1.}\em
Let $ \mathcal{X} = \{\x| \x \in \{0, 1\}^{d}\}. $ Let $ \mathcal{F} $ be the class of linear predictors, i.e.,
\begin{equation}
\mathcal{F} = \{\b\Tr\x | \b \in \mathbb{R}^{d}\, and\, \norm{\b}_{1} \le B\}.
\end{equation}
We have 
\begin{equation}
\hat{\mathfrak{R}}_{s}(\mathcal{F}) \le B\sqrt{\dfrac{2\ln\left(d\right)}{n}} \label{rf}
\end{equation}
\em
\begin{proof}
Let $ \mathcal{S} = \{\x^{(1)}\dots \x^{(n)}\} $ be a data set of $ n $ samples. Denote $ x_{j} $ as the j-th element of $ \x. $
\begin{align}
\hat{\mathfrak{R}}_{s}(\mathcal{F}) &= \mathbb{E}_{\sigma}\left[\sup_{f \in \mathcal{F}} \left(\dfrac{1}{n}\sum\limits_{i=1}^{n}\sigma_{i}f(\x^{(i)})\right)\right]\\
& = \dfrac{1}{n}\mathbb{E}_{\sigma}\left[\sup_{f \in \mathcal{F}} \left(\sum\limits_{i=1}^{n}\sigma_{i}\b\Tr \x^{(i)}\right)\right]\\
& = \dfrac{1}{n}\mathbb{E}_{\sigma}\left[\sup_{\b: \norm{\b}_{1}\le B} \left(\b\Tr\left(\sum\limits_{i=1}^{n}\sigma_{i} \x^{(i)}\right)\right)\right]\label{b}\\
& = \dfrac{B}{n} \mathbb{E}_{\sigma}\left[\norm{\sum\limits_{i=1}^{n}\sigma_{i}\x^{(i)}}_{\infty}\right] \label{holder}\\
& = \dfrac{B}{n} \mathbb{E}_{\sigma}\left[\sup_{j \in \{1\dots d\}}\left(\sum\limits_{i=1}^{n}\sigma_{i}\x^{(i)}\right)_{j}\right] \label{k+1}\\
& \le \dfrac{B\sqrt{2\ln\left(d\right)}}{n} \sup_{j \in \{1\dots d\}}\sqrt{\sum\limits_{i=1}^{n}\left[x^{(i)}_{j}\right]^{2}} \label{massart}\\
& \le \dfrac{B\sqrt{2\ln\left(d\right)}}{n}\sqrt{n \norm{x}^{2}_{\infty}}\\
& = \norm{\x}_{\infty}B\sqrt{\dfrac{2\ln\left(d\right)}{n}}\\
& = B\sqrt{\dfrac{2\ln\left(d\right)}{n}} \label{x}
\end{align}
Equation \ref{holder} uses Holder's inequality when the equal sign is taken. inequality \ref{massart} uses Massart's finite class lemma. Equation \ref{x} is from the fact that $ \x \in \{0, 1\}^{d}. $ Therefore we proved inequality \ref{rf}.
\end{proof}

\paragraph{Remark 1.} \em
Function
\begin{equation}
\phi(g) = \ln\left(1 + \exp(g)\right)
\end{equation} 
is 1-Lipschitz continuous for $ g \in \mathbb{R}. $
\em
\begin{proof}
$ \abs{\partial \phi(g)/\partial g} = Sigmoid(g) \le 1. $
\end{proof}

\paragraph{Lemma 2.}\em
Let $ \mathcal{X} = \{\x| \x \in \{0, 1\}^{d}\},  \mathcal{F} $ be a class of linear predictors, i.e.,
\begin{equation}
\mathcal{F} = \{\b\Tr\x | \b \in \mathbb{R}^{d}\}.
\end{equation}
Let $ \mathcal{G} $ be another class of linear predictors, i.e.,
\begin{equation}
\mathcal{G} = \{\w\Tr\x+c | \w \in \mathbb{R}^{d}, c \in \mathbb{R}\}.
\end{equation}
Let $ \mathcal{H} $ be a function of $ \mathcal{F} $ and $ \mathcal{G} $, written as 
\begin{equation}
\mathcal{H} = \{\ln\left[\exp\left(f(\x)\right) + \exp\left(f(\x)+g(\x)\right)\right] | f \in \mathcal{F}, g \in \mathcal{G}\},
\end{equation}
Let $ \mathcal{S} = \{\x^{(1)}\dots \x^{(n)}\} $ be a data set of $ n $ samples. We have 
\begin{equation}
\hat{\mathfrak{R}}_{s}(\mathcal{H}) \le \hat{\mathfrak{R}}_{s}(\mathcal{F}) + \hat{\mathfrak{R}}_{s}(\mathcal{G}) \label{hfg}
\end{equation}
\em
\begin{proof}
\begin{align}
\hat{\mathfrak{R}}_{s}(\mathcal{H}) &=  \mathbb{E}_{\sigma}\left[\sup_{h \in \mathcal{H}} \left(\dfrac{1}{n}\sum\limits_{i=1}^{n}\sigma_{i}h(\x^{(i)})\right)\right]\\
&= \mathbb{E}_{\sigma}\left[\sup_{f \in \mathcal{F}, g \in \mathcal{G}} \left(\dfrac{1}{n}\sum\limits_{i=1}^{n}\sigma_{i}\ln\left[\exp\left(f(\x^{i})\right) + \exp\left(f(\x^{i})+g(\x^{i})\right)\right]\right)\right]\\
&= \mathbb{E}_{\sigma}\left[\sup_{f \in \mathcal{F}, g \in \mathcal{G}} \left(\dfrac{1}{n}\sum\limits_{i=1}^{n}\sigma_{i}\ln\left[\exp\left(f(\x^{i})\right)\right] +\ln\left[1+ \exp\left(g(\x^{i})\right)\right]\right)\right]\\
&\le \mathbb{E}_{\sigma}\left[\sup_{f \in \mathcal{F}} \left(\dfrac{1}{n}\sum\limits_{i=1}^{n}\sigma_{i}f(\x^{i})\right)\right] + \mathbb{E}_{\sigma}\left[\sup_{g \in \mathcal{G}}\left(\dfrac{1}{n}\sum\limits_{i=1}^{n}\sigma_{i}\ln\left[1+ \exp\left(g(\x^{i})\right)\right]\right)\right] \label{twofunc}\\
&= \mathbb{E}_{\sigma}\left[\sup_{f \in \mathcal{F}} \left(\dfrac{1}{n}\sum\limits_{i=1}^{n}\sigma_{i}f(\x^{i})\right)\right] + \mathbb{E}_{\sigma}\left[\sup_{g \in \mathcal{G}}\left(\dfrac{1}{n}\sum\limits_{i=1}^{n}\sigma_{i}\phi(g(\x^{(i)}))\right)\right] \\
& \le \hat{\mathfrak{R}}_{s}(\mathcal{F}) + \hat{\mathfrak{R}}_{s}(\mathcal{G})
\end{align}
\end{proof}
In inequality \ref{twofunc}, the first part of it is exactly the Rademacher complexity of $ \mathcal{F} $ by definition. The second part can be shown to be $ \le \hat{\mathfrak{R}}_{s}(\mathcal{G}) $ by using Ledoux-Talagrand Contraction Lemma, combining with the results in \textbf{Remark 1} that $ \phi(g) $ is 1-Lipschitz continuous. Hence we proved inequality \ref{hfg}.

\paragraph{Remark 2.} \em
Let $ \mathcal{X} = \{\x| \x \in \{0, 1\}^{d}\}. $ Let $ \mathcal{G} $ be the class of linear predictors, i.e.,
\begin{equation}
\mathcal{G} = \{\w\Tr\x+c | \w \in \mathbb{R}^{d}, c \in \mathbb{R}\, and\, \norm{\w}_{1}\le W\},
\end{equation}
where $ \W_{j} $ is the j-th column of $ \W. $ Let $ \mathcal{S} = \{\x^{(1)}\dots \x^{(n)}\} $ be a data set of $ n $ samples. We have 
\begin{equation}
\hat{\mathfrak{R}}_{s}(\mathcal{G}) \le W\sqrt{\dfrac{2\ln\left(d\right)}{n}} \label{rg}
\end{equation}
\em
\begin{proof}
\begin{align}
\hat{\mathfrak{R}}_{s}(\mathcal{G}) &= \mathbb{E}_{\sigma}\left[\sup_{g \in \mathcal{G}} \left(\dfrac{1}{n}\sum\limits_{i=1}^{n}\sigma_{i}g(\x^{(i)})\right)\right]\\
&= \mathbb{E}_{\sigma}\left[\sup_{g \in \mathcal{G}} \left(\dfrac{1}{n}\sum\limits_{i=1}^{n}\sigma_{i}(\w\Tr\x^{(i)}+c)\right)\right]\\
& \le \dfrac{1}{n}\mathbb{E}_{\sigma}\left[\sup_{\w: \norm{\w}_{1}\le W} \left(\w\Tr\left(\sum\limits_{i=1}^{n}\sigma_{i}\x^{(i)}\right)\right)+\sup_{c} \left(\sum\limits_{i=1}^{n}\sigma_{i}\right)\right]\\
& = \dfrac{1}{n}\mathbb{E}_{\sigma}\left[\sup_{\w: \norm{\w}_{1}\le W} \left(\w\Tr\left(\sum\limits_{i=1}^{n}\sigma_{i}\x^{(i)}\right)\right)\right]+\dfrac{1}{n}\mathbb{E}_{\sigma}\left[\sup_{c} \left(\sum\limits_{i=1}^{n}\sigma_{i}\right)\right]\label{two}\\
\end{align}
Notice that the first part of equation \ref{two} can be bounded by $ W\sqrt{\dfrac{2\ln\left(d\right)}{n}} $ by using the results in \textbf{Lemma 1}, and the second part is exactly $ 0 $ by the definition of Rademacher complexity. Thus we proved inequality \ref{rg}.
\end{proof}

\paragraph{Theorem 1.}\em
Let $ \mathcal{X} = \{\x| \x \in \{0, 1\}^{k}\}, \mathcal{S} = \{\x^{(1)}\dots \x^{(n)}\} $ be a data set of $ n $ samples. Given a restricted Boltzmann machine with $ k $ visible units and $ m $ hidden ones. For all the parameters $ \btheta = \{\c, \b, \W\}, \c \in \mathbb{R}^{m}, \b \in \mathbb{R}^{k}, $ and $ \W \in \mathbb{R}^{k \times m}, $ assuming $ \b, \W $ are bounded by spheres $ \norm{\b}_{1} \le B, \norm{\W}_{\max} =\forall j\, \norm{\W_{j}}_{1}\le W,$ where $ \W_{j} $ is the j-th column of $ \W. $ We can bound the empirical Rademacher complexity for the likelihood of this restricted Boltzmann machine as:
\begin{equation}
\hat{\mathfrak{R}}_{s}(\ln p_{\btheta}) \le m\sqrt{\dfrac{2\ln\left(k\right)}{n}}\left(B + W\right). \label{final}
\end{equation}
\em
\begin{proof}
As we stated, we only consider equation \ref{decoupled} that has randomness and ignore the partition part of the log-likelihood. Using the notation in \textbf{Lemma 2}, we can write 
\begin{equation}
\hat{\mathfrak{R}}_{s}(\ln p_{\btheta}) = \hat{\mathfrak{R}}_{s}(\sum\limits_{j=1}^{m}\mathcal{H}_{j}) \le \sum\limits_{j=1}^{m}\hat{\mathfrak{R}}_{s}(\mathcal{H}_{j}), \label{ele}
\end{equation} 
which is from the elementary properties of the Rademacher complexity. Knowing the fact that each $ \mathcal{H}_{j} $ is of the same  hypothesis space further provides us with 
\begin{equation}
\sum\limits_{j=1}^{m}\hat{\mathfrak{R}}_{s}(\mathcal{H}_{j}) = m\hat{\mathfrak{R}}_{s}(\mathcal{H}). \label{mh}
\end{equation}   
From \textbf{Lemma 2} we have $ \hat{\mathfrak{R}}_{s}(\mathcal{H}) \le \hat{\mathfrak{R}}_{s}(\mathcal{F}) + \hat{\mathfrak{R}}_{s}(\mathcal{G}). $ And from \textbf{Lemma 1} we can directly bound $ \hat{\mathfrak{R}}_{s}(\mathcal{F}) $ by $ B\sqrt{\dfrac{2\ln\left(k\right)}{n}}. $ For $ \hat{\mathfrak{R}}_{s}(\mathcal{G}), $ by using the results in \textbf{Remark 2}, we can bound it by $ W\sqrt{\dfrac{2\ln\left(k\right)}{n}}. $ Together with inequality \ref{ele} and equation \ref{mh}, we proved this theorem.
\end{proof}

\section{Rademacher Complexity with CD-1 Approximation}
Contrastive Divergence is an approximation of the log-likelihood gradient that has been found to be a successful update rule for training RBMs. The reason that we are applying contrastive divergence algorithm is that because the partition function is can be hardly estimated by enumerating all the possible values because the complexity will be in the order of exponential, nor the factorization trick we used for the numerator can be used. In order to approximate the partition function for all possible visible examples, a MCMC chain is created. First an example is sampled uniformly from the empirical training examples. Then a mean-field approximation is applied to obtain the values of hidden units (whose values are also binary): Rather than sample from the distribution of $ \h, $ we use the values $ \forall\, i, P(h_{i}= 1)  $ as the values to approximate the samples. After we obtain $ \tilde{\h} $, we have the distribution of $ \x $ based on the current values of $ \h $ (mean-field approximation) and parameters ($ \W, \b, \c $). We sample from this distribution to obtain a vector $ \tilde{\x} $ and use it to approximate the partition function. This procedure can also extended to more steps (CD-k, k steps). But experiments have shown that, even one step (CD-1) can yield a good performance for the model \cite[]{Bengio2009}.

After using CD-1 algorithm, the Rademacher complexity of the second part of equation \ref{LL} is no more free of randomness, due to the fact that $ \tilde{\x} $ is a function of $ \x. $ If we rewrite the second part as 
\begin{equation}
Z_{\btheta} \approx \sum\limits_{\h}\exp\{-Energy(\tilde{\x}, \h; \btheta)\}, \label{appro}
\end{equation}
Rademacher complexity of this term is also depending on random variable $ \x. $

To simplify the procedure but without losing generality, instead of sampling $ \tilde{\x} $ from its distribution, we also use mean field approximation to obtain $ \tilde{\x}. $ Also, we can write the energy function as 
\begin{equation}
Energy(\x, \h; \btheta) = \x\Tr\W\h,
\end{equation}
while ignore the bias term for simplicity.

\paragraph{Remark 3.} \em
Using mean filed approximation, we can obtain 
\begin{equation}
\tilde{\h}\Tr = ( sgm(\x\Tr\W_{\cdot 1}), \dots,  sgm(\x\Tr\W_{\cdot m}) ),
\end{equation}
where sgm() is the sigmoid function, and $ \W_{\cdot j} $ is the $ j $-th column of $ \W. $
\em
\begin{proof}
$ P(\tilde{\h}|_{\x} = \1) = \dfrac{\exp \{ \x\Tr\W\1\}}{\sum\limits_{\h}\exp \{ \x\Tr\W\h\}}=\dfrac{\prod\limits_{i=1}^{m}\exp \{ \x\Tr\W_{\cdot i}\}}{\prod\limits_{i=1}^{m}\sum\limits_{h_{i}}\exp \{ \x\Tr\W_{\cdot i} h_{i}\}}$\\
With the fact that $ \forall i,j\,  h_{i} \perp h_{j} |_{\x},  \forall i, P(\tilde{h_{i}}|_{\x} = 1) = \dfrac{\exp \{ \x\Tr\W_{\cdot i}\}}{1 + \exp \{ \x\Tr\W_{\cdot i}\}} = sgm(\x\Tr\W_{\cdot i}). $
\end{proof}

\paragraph{Remark 4.} \em
Similar to \textbf{Remark 3}, we can obtain  
\begin{equation}
\tilde{\x}\Tr = ( sgm(\W_{1\cdot}\tilde{\h}), \dots, sgm(\W_{k\cdot}\tilde{\h}) ), 
\end{equation}
where $ \forall v\, \W_{v\cdot} $ is the $ v $-th row of $ \W. $
\em

\paragraph{Lemma 3.} \em
By using CD-1 Algorithm, and mean field approximation for both $ \tilde{\x} $ and $ \tilde{\h}, $ we have part 2 of equation \ref{LL} as 
\begin{equation}
\ln Z_{\theta} = \sum\limits_{j=1}^{m}\ln\left[ 1+\exp \{ \sum\limits_{i=1}^{k} \W_{ij}sgm\left( \sum\limits_{v=1}^{m}\W_{iv}sgm(\x\Tr\W_{\cdot v}) \right) \} \right],
\end{equation}
where we use $ \W_{ij} $ to denote the element of $ i $-th row and $ j $-th column of matrix $ \w. $
\em
\begin{proof}
\textit{(sketch)} Using the results from \textbf{Remark 3} and \textbf{Remark 4}, and the same factorization trick used before in equation \ref{fac}, this can be shown easily.
\end{proof}

\paragraph{Lemma 4.}\em
Let $ \mathcal{X} = \{\x| \x \in \{0, 1\}^{k}\}$,  Let $ \mathcal{T} $ be a compositional function of $ \x $ with parameters $ \W $, i.e.,
\begin{equation}
\mathcal{T} = \{ \W_{uj}sgm\left( \sum\limits_{v=1}^{m}\W_{uv}sgm(\x\Tr\W_{\cdot v}) \right) | \W \in \mathbb{R}^{k \times m},\, \forall u \in \{1,\dots, k\}, \forall j \in \{1,\dots, m\} \},
\end{equation} 
and assuming $\W $ is bounded by spheres $ \norm{\W}_{\max} =\forall j\, \norm{\W_{\cdot j}}_{1}\le W,$ where $ \W_{\cdot j} $ is the j-th column of $ \W. $ Also Let $ \mathcal{S} = \{\x^{(1)}\dots \x^{(n)}\} $ be a data set of $ n $ samples. We have 
\begin{equation}
\hat{\mathfrak{R}}_{s}(\mathcal{T}) \le \dfrac{W\sqrt{2n\ln |\mathcal{T}|}}{n}
\end{equation}
\em
\begin{proof}
\begin{align}
\hat{\mathfrak{R}}_{s}(\mathcal{T}) &=  \mathbb{E}_{\sigma}\left[\sup_{t_{\w} \in \mathcal{T}} \left(\dfrac{1}{n}\sum\limits_{i=1}^{n}\sigma_{i}t_{\W}(\x^{(i)})\right)\right] \Longrightarrow\\
\exp\{ \mathbb{E}_{\sigma}\left[s\sup_{t_{\w} \in \mathcal{T}} \left(\sum\limits_{i=1}^{n}\sigma_{i}t_{\W}(\x^{(i)})\right)\right] \} &\le \mathbb{E}_{\sigma}\left[ \exp\{ s\sup_{t_{\w} \in \mathcal{T}} \left(\sum\limits_{i=1}^{n}\sigma_{i}t_{\W}(\x^{(i)})\right) \} \right] \label{jensen}\\
&= \sup_{t_{\w} \in \mathcal{T}}\mathbb{E}_{\sigma}\left[ \exp\{ s \left(\sum\limits_{i=1}^{n}\sigma_{i}t_{\W}(\x^{(i)})\right) \} \right]\\
&\le \sum\limits_{t_{\w} \in \mathcal{T}}\mathbb{E}_{\sigma}\left[ \exp\{ s \left(\sum\limits_{i=1}^{n}\sigma_{i}t_{\W}(\x^{(i)})\right) \} \right]\\
&= \sum\limits_{t_{\w} \in \mathcal{T}}\mathbb{E}_{\sigma}\left[ \prod\limits_{i=1}^{n}\exp\{ s \left(\sigma_{i}t_{\W}(\x^{(i)})\right) \} \right]\\
&= \sum\limits_{t_{\w} \in \mathcal{T}}\prod\limits_{i=1}^{n}\mathbb{E}_{\sigma_{i}}\left[ \exp\{ s \left(\sigma_{i}t_{\W}(\x^{(i)})\right) \} \right]\label{ind}\\
&\le \sum\limits_{t_{\w} \in \mathcal{T}}\prod\limits_{i=1}^{n} \exp\{ \dfrac{4s^{2}\W_{uj}^{2}}{8} \}  \label{hoeffding}\\
&= \sum\limits_{t_{\w} \in \mathcal{T}} \exp\{ \dfrac{4ns^{2}\W_{uj}^{2}}{8} \}\\
&\le |\mathcal{T}|\sup_{t_{\w} \in \mathcal{T}}\exp\{ \dfrac{4ns^{2}\W_{uj}^{2}}{8} \}\label{bound}\\
&= |\mathcal{T}|\exp\{ \dfrac{ns^{2}W^{2}}{2} \}\Longrightarrow\\
\exp\{ \mathbb{E}_{\sigma}\left[s\sup_{t_{\w} \in \mathcal{T}} \left(\sum\limits_{i=1}^{n}\sigma_{i}t_{\W}(\x^{(i)})\right)\right] \} &\le \dfrac{ \ln|\mathcal{T}|}{s} + \dfrac{nsW^{2}}{2}\Longrightarrow \label{beforederi}\\
\exp\{ \mathbb{E}_{\sigma}\left[s\sup_{t_{\w} \in \mathcal{T}} \left(\sum\limits_{i=1}^{n}\sigma_{i}t_{\W}(\x^{(i)})\right)\right] \} &\le W\sqrt{2n\ln |\mathcal{T}|} \label{afterderi}\Longrightarrow\\
\hat{\mathfrak{R}}_{s}(\mathcal{T}) =  \mathbb{E}_{\sigma}\left[\sup_{t_{\w} \in \mathcal{T}} \left(\dfrac{1}{n}\sum\limits_{i=1}^{n}\sigma_{i}t_{\W}(\x^{(i)})\right)\right] &\le \dfrac{W\sqrt{2n\ln |\mathcal{T}|}}{n} \label{last}
\end{align}
\end{proof}
Inequality \ref{jensen} uses Jensen's inequality, and equation \ref{ind} uses the independence property of expectation. To obtain \ref{hoeffding}, we first notice that $ sgm() \in (0,1), $ thus $ t_{\w}(\x_{i}) \in (0, \W_{uj}) $ and $ \sigma_{i}t_{\w}(\x_{i}) \in (-\W_{uj}, \W_{uj}), $ and then use Hoeffding's Inequality. Inequality \ref{bound} uses our assumption that $ \norm{\W}_{\max} \le W. $ By taking derivative of the RHS of \ref{beforederi} and set it to $ 0, $ we obtained $ s = \sqrt{\dfrac{2\ln |\mathcal{T}|}{nW}} $ hence we obtain equation \ref{afterderi}. By dividing both sides by $ n $ we obtain equation \ref{last}.

\paragraph{Corollary 1.}\em
Let $ \mathcal{X} = \{\x| \x \in \{0, 1\}^{k}\}, \mathcal{S} = \{\x^{(1)}\dots \x^{(n)}\} $ be a data set of $ n $ samples. Given a restricted Boltzmann machine with $ k $ visible units and $ m $ hidden ones, and it is trained by using CD-1 algorithm. For all the parameters $ \btheta = \{\W\}, $ and $ \W \in \mathbb{R}^{k \times m}, $ assuming $\W $ is bounded by spheres $ \norm{\W}_{\max} =\forall j\, \norm{\W_{\cdot j}}_{1}\le W,$ where $ \W_{\cdot j} $ is the j-th column of $ \W. $ Let $ \mathcal{T} $ be a compositional function of $ \x $ with parameters $ \W $, i.e.,
\begin{equation}
\mathcal{T} = \{ \W_{ij}sgm\left( \sum\limits_{v=1}^{m}\W_{iv}sgm(\x\Tr\W_{\cdot v}) \right) | \W \in \mathbb{R}^{k \times m}\, \forall j \in \{1,\dots, m\}.
\end{equation} 
We further assume the VC-dimension of $ \mathcal{T} $ is $ VC(\mathcal{T}). $  We can bound the empirical Rademacher complexity for the likelihood of this restricted Boltzmann machine as:
\begin{equation}
\hat{\mathfrak{R}}_{s}(\ln p_{\btheta}) \le \dfrac{W}{\sqrt{n}}\left(m\sqrt{2 \ln k} + k\sqrt{2VC(\mathcal{T}) \ln (n+1)} \right). \label{final2}
\end{equation}
\em
\begin{proof}
Using the result of \textbf{Remark 1}, we can bound $\hat{\mathfrak{R}}_{s}(\log Z_{\theta})$ in equation \ref{appro} by $ \hat{\mathfrak{R}}_{s}(\log Z_{\theta}) \le \hat{\mathfrak{R}}_{s}(\sum\limits_{i=1}^{k}\mathcal{T}_{i}). $ Similar to equation \ref{ele}, $ \hat{\mathfrak{R}}_{s}(\sum\limits_{i=1}^{k}\mathcal{T}_{i}) \le \sum\limits_{i=1}^{k}\hat{\mathfrak{R}}_{s}(\mathcal{T}_{i}).$ Knowing the fact that each $ \mathcal{T}_{i} $ is from the same hypothesis space further provides us with 
\begin{equation}
\sum\limits_{i=1}^{k}\hat{\mathfrak{R}}_{s}(\mathcal{T}_{i}) = k\hat{\mathfrak{R}}_{s}(\mathcal{T}).
\end{equation}
Using the results in \textbf{Lemma 4} we obtain $\hat{\mathfrak{R}}_{s}(\log Z_{\theta}) \le \dfrac{Wk\sqrt{2n\ln |\mathcal{T}|}}{n}. $ Then by Sauer-Shelah lemma, we know
\begin{equation}
\max\limits_{\mathcal{S}}|\mathcal{T(\mathcal{S})}| \le (n+1)^{VC(\mathcal{T})}.
\end{equation}
Therefore we obtain
\begin{equation}
\hat{\mathfrak{R}}_{s}(\log Z_{\theta}) \le \dfrac{Wk\sqrt{2n\ln |\mathcal{T}|}}{n} \le  \dfrac{Wk\sqrt{2VC(\mathcal{T})n\ln (n+1)}}{n}
\end{equation}
Together with the results from \textbf{Theorem 1}, while ignoring the bias term, we proved this corollary.
\end{proof}

\section{Future Direction}
Can we get a tighter bound on it? Can we extend this results to multi-layer Boltzmann machines, like deep belief networks (DBN) or deep Boltzmann machines (DBM)? Is that possible to obtain the exact expression of the VC dimension of our constructed function $ \mathcal{T} $?

%\nocite{*}
\bibliographystyle{plainnat}
\bibliography{all}

\begin{thebibliography}{8}
\providecommand{\natexlab}[1]{#1}
\providecommand{\url}[1]{\texttt{#1}}
\expandafter\ifx\csname urlstyle\endcsname\relax
  \providecommand{\doi}[1]{doi: #1}\else
  \providecommand{\doi}{doi: \begingroup \urlstyle{rm}\Url}\fi

\bibitem[Bengio(2009)]{Bengio2009}
Yoshua Bengio.
\newblock Learning deep architectures for ai.
\newblock \emph{Found. Trends Mach. Learn.}, 2\penalty0 (1):\penalty0 1--127,
  January 2009.
\newblock ISSN 1935-8237.

\bibitem[Hinton and Salakhutdinov(2012)]{Hinton2012}
Geoffrey Hinton and Ruslan Salakhutdinov.
\newblock {A better way to pretrain deep Boltzmann machines}.
\newblock \emph{Advances in Neural Information}, \penalty0 (3):\penalty0 1--9,
  2012.

\bibitem[Hinton and Salakhutdinov(2006)]{Hinton2006a}
Geoffrey~E. Hinton and R~R Salakhutdinov.
\newblock {Reducing the dimensionality of data with neural networks.}
\newblock \emph{Science (New York, N.Y.)}, 313\penalty0 (5786):\penalty0
  504--7, 2006.

\bibitem[Hinton et~al.(2006)Hinton, Osindero, and Teh]{Hinton2006b}
Geoffrey~E. Hinton, Simon Osindero, and Yee~Whye Teh.
\newblock {A fast learning algorithm for deep belief nets.}
\newblock \emph{Neural computation}, 18\penalty0 (7):\penalty0 1527--54, 2006.

\bibitem[Honorio(2012)]{honorio2012}
Jean Honorio.
\newblock {Lipschitz parametrization of probabilistic graphical models}.
\newblock \emph{Uncertainty in Artificial Intelligence 2012}, 2012.

\bibitem[Salakhutdinov and Hinton(2009)]{Salakhutdinov2009}
Ruslan Salakhutdinov and Geoffrey Hinton.
\newblock {Deep Boltzmann Machines}.
\newblock \emph{Artificial Intelligence}, 5\penalty0 (2):\penalty0 448--455,
  2009.

\bibitem[Salakhutdinov and Hinton(2012)]{Salakhutdinov2012}
Ruslan Salakhutdinov and Geoffrey Hinton.
\newblock {An Efficient Learning Procedure for Deep Boltzmann Machines}.
\newblock \emph{Neural Computation}, 24\penalty0 (8):\penalty0 1967--2006,
  2012.

\bibitem[Smolensky(1986)]{Smolensky1986}
Paul Smolensky.
\newblock {Information processing in dynamical systems: Foundations of harmony
  theory}.
\newblock \emph{Parallel Distributed Processing Explorations in the
  Microstructure of Cognition}, 1\penalty0 (1):\penalty0 194--281, 1986.

\end{thebibliography}

\end{document}